\documentclass[11pt]{article}

\usepackage[final]{acl}

\usepackage{times}
\usepackage{latexsym}

\usepackage[T1]{fontenc}

\usepackage[utf8]{inputenc}
\usepackage[table]{xcolor}
\usepackage{colortbl}

\usepackage{microtype}

\usepackage{inconsolata}

\usepackage{graphicx}

\usepackage{algorithm}
\usepackage{algorithmic}
\usepackage{amsmath}
\usepackage{booktabs}
\usepackage{multirow}
\usepackage{subfigure}
\usepackage{tabularx}
 \newcolumntype{Y}{>{\centering\arraybackslash}X}
\newcommand{\Lo}{\textsuperscript{\scriptsize\textsc{l}}}
\newcommand{\Hi}{\textsuperscript{\scriptsize\textsc{h}}}

\usepackage{makecell}
\usepackage{array}
\usepackage[most]{tcolorbox}
\newtcolorbox{prompt}[1]{
  colback=gray!10,
  colframe=gray!80!black,
  fonttitle=\bfseries,
  title={#1}
}

\title{ENPMR-Bench: Benchmarking Proactive Memory Retrieval for Emotional Support Agents}

\author{
    \textbf{
    Xing Fu,
    Yulin Hu,
    Mengtong Ji,
    Haozhen Li,
    Yixin Sun
    }\\
    \textbf{
    Weixiang Zhao,
    Yanyan Zhao\thanks{Corresponding author.},
    Bing Qin
    }\\
    Research Center for Social Computing and Interactive Robotics \\
    Harbin Institute of Technology, Heilongjiang, China \\
    \texttt{\{xfu, yyzhao\}@ir.hit.edu.cn}
}

\begin{document}
\maketitle

\begin{abstract}

Memory-augmented language agents are increasingly deployed in affective applications such as emotional support, where understanding and responding to users' latent emotional needs is critical. However, existing research often treats memory as a tool for factual retrieval, overlooking its role in shaping users' emotional experiences. In this work, we introduce ENPMR-Bench, a benchmark for evaluating Emotional Need-aware Proactive Memory Retrieval (ENPMR), a core capability that enables agents to infer users' latent emotional needs and proactively retrieve appropriate memories to support empathetic interaction. Grounded in Maslow's hierarchy of needs, ENPMR-Bench includes over 1,800 memory-augmented dialogues and defines structured mappings between emotional needs and supportive memory types. Experimental results demonstrate that current retrieval paradigms, including both embedding-based and LLM-driven approaches, exhibit substantial deficiencies, with empathy scores significantly lagging behind golden memory conditions. While chain-of-thought prompting improves the alignment between inferred emotional needs and retrieved memories to some extent, a notable performance gap remains. Together, these findings reveal critical limitations in current agents and outline directions for advancing personalized emotional support through need-sensitive memory retrieval.\footnote{Our data and code are available at \url{https://github.com/xingqwq/ENPMR-Bench}}

\end{abstract}

\section{Introduction}

\begin{figure}[ht!]
\centerline{\includegraphics[width=\linewidth]{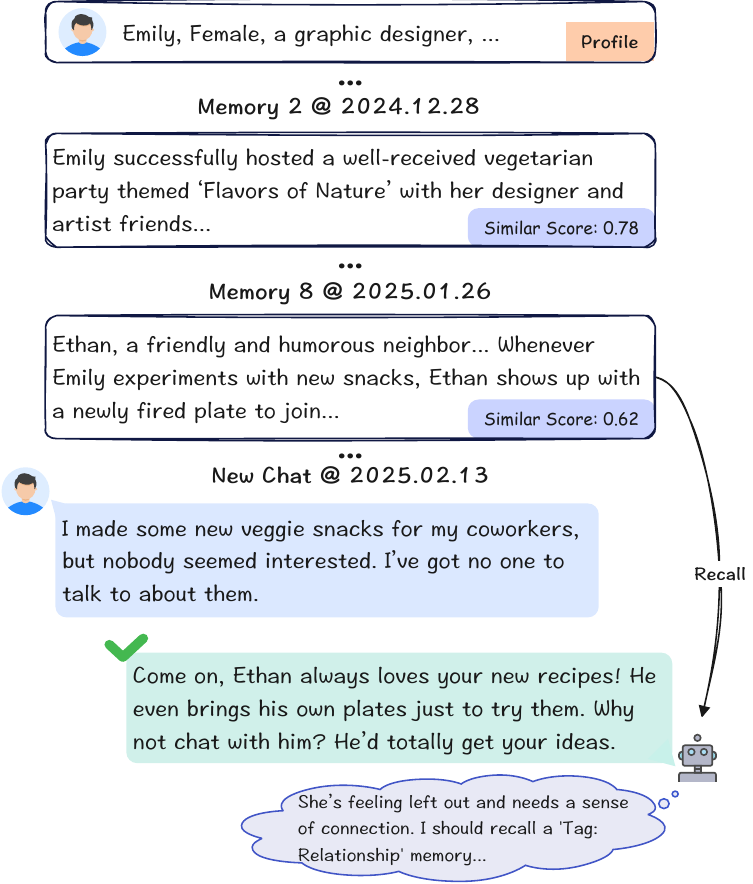}}
\caption{An example of emotional need-aware proactive memory retrieval, where the agent infers a latent need from social disconnection and retrieves an interpersonal memory to foster connection.}
\label{fig:intro_fig}
\end{figure}

Recently, LLM–based agents have been increasingly integrated into users' daily lives~\cite{hua2023tutorial, meyer2024you, wang2026care}, with a growing number of users relying on them in affective scenarios such as emotional support~\cite{phang2025investigating}. As these applications evolve, memory has been widely recognized as a foundational mechanism for enabling long-term interaction and personalization~\cite{zheng2025lifelong, yehudai2025survey}. However, existing research often conceptualizes memory primarily as a mechanism for factual retention~\citep{huang2024emotional}, neglecting how retrieved memories influence users’ emotional experiences.

In emotional support scenarios, memory usage differs from task-driven dialogues.~\citep{zhang2016combining}. Users often express their emotions in an indirect and nuanced manner, rather than explicitly articulating their emotional needs~\citep{weger2014relative, socio-emotional}. Consequently, as illustrated in Figure~\ref{fig:intro_fig}, to provide effective support, agents are required to infer users' latent emotional needs~\citep{zhao2022cauain} and proactively retrieve memories that are emotionally salient and supportive, instead of relying solely on surface-level semantic similarity~\citep{scherer2005emotions, yu2025more}. Moreover, psychological studies indicate that in long-term emotional interactions, both neglecting past experiences and invoking inappropriate memories can undermine trust and emotional coherence.~\citep{kasap2009making, leite2013social, Wolf2024How, wang2025psychological}. \emph{Therefore, the ability to infer emotional needs and retrieve appropriate memories constitutes a core component of emotional intelligence, which we define as emotional need-aware proactive memory retrieval} (ENPMR).

Despite recent progress in memory-augmented agents~\cite{yehudai2025survey,zhao2025llms, lin2024interpretable, zhao2025teaching}, this capability has not yet been systematically evaluated. Most memory benchmarks~\cite{locomo, longmemeval, perltqa} adopt a QA paradigm that presumes users can explicitly express their emotional needs and specify what information should be retrieved. However, as shown in Figure~\ref{fig:intro_fig}, this assumption collapses in emotional support settings. And, existing benchmarks typically rely on static or single-dimensional user modeling strategies. They fail to capture comprehensive representations of user attributes such as life goals or significant personal experiences, which are critical for tailoring emotionally resonant responses~\citep{king2001health}. As a result, agents struggle to utilize diverse memory resources, limiting the scope for evaluating.

To fill the gap, we propose ENPMR-Bench, a benchmark designed to systematically evaluate the capacity of ENPMR. Grounded in Maslow’s hierarchy of needs~\cite{maslow} and expert-designed retrieval principles, the benchmark defines structured mappings between emotional needs and appropriate memory types. Based on this framework, we construct over 1,800 memory-augmented emotional support dialogues spanning diverse themes and systematically evaluate embedding-based retrieval, agentic memory systems, and LLMs.

Our results show that retrieval deficiencies substantially degrade downstream empathetic performance. Even the best model achieves only 46.41\% Recall@10, and all fall below 10\% in Top-1. Under such retrieved memories, DeepSeek-v3 scores 4.38 in empathy, improving to 4.91 with golden memories, underscoring the importance of memory appropriateness. Meanwhile, chain-of-thought(CoT) reasoning brings moderate gains but a considerable gap remains compared to the upper bound, indicating persistent limitations in agents' ability to select emotionally appropriate memories. Our main contributions are as follows:

\begin{itemize}
    \item We establish a comprehensive evaluation protocol for proactive memory retrieval in emotional support scenario.
    \item We present a systematic comparison of representative agents' memory systems and LLMs, revealing their strengths and limitations.
    \item We identify key challenges for future research and discuss directions to better facilitate personalized emotional support.
\end{itemize}

\section{Related Works}

\subsection{Memory Benchmark}

Existing memory benchmarks mainly evaluate agents’ ability to retrieve factual knowledge. For long-document retrieval, prior work evaluates whether agents can extract relevant information from large-scale corpora~\cite{kuratov2024search, zhang2024bench, liu2023lost}. In multi-session dialogue settings, \citet{locomo} introduce a long-term dialogue corpus with temporally structured events. \citet{longmemeval} focus on information extraction and reasoning in ultra-long, timestamped user–assistant interactions. \citet{perltqa} construct long-term dialogues grounded in character biographies and life events. \citet{membench} further evaluate memory usage across participatory and observational settings with both factual and reflective memories. Despite the coverage of various task types, most existing benchmarks mainly rely on QA pairs with predefined retrieval targets. Such a passive retrieval paradigm may not adequately reflect agent capabilities in open-domain scenarios. Recent work explores emotion in memory retrieval~\cite{memdial}, but still overlooks the alignment between user needs and memory types, which is essential for emotional support.

\begin{table*}[ht!]
    \renewcommand{\arraystretch}{1.25}
    \centering
    \resizebox{\linewidth}{!}{
        \begin{tabular}{
        >{\centering\arraybackslash}m{4.5cm}
        >{\raggedright\arraybackslash}m{7.5cm}
        >{\centering\arraybackslash}m{5.5cm}
        }
        \toprule
        \textbf{Users' Emotional Needs} & \textbf{Manifestation of Deficiency} & \textbf{Types of Memories Recalled} \\
        \midrule
        Physiological Needs(PN) & 
        The user feels fatigue, hunger, or other forms of physical discomfort. & 
        Preference \\
        
        \midrule
        {Love and Belonging(LB)} & 
        The user tends to feel lonely, rejected, or lacks a sense of acceptance and belonging. & 
        Relationship \\
        
        \midrule
        {Esteem Needs(EN)} & 
        The user often shows signs of inferiority, lack of self-confidence, and is highly sensitive to external evaluation. & 
        Highlight, Relationship \\
        
        \midrule
        {Self-actualization Needs(SA)} & 
        The user feels lost, lacks a sense of purpose or passion, doubts their own value, and shows little interest in life or work. & 
        Goal, Power, Highlight \\
        
        \bottomrule
        \end{tabular}
    }
    \caption{The mapping between unmet user needs and corresponding memory recall types.}
    \label{tab:need_memory_recall}
\end{table*}

\subsection{Emotional Support Dialogue}

Early work on emotional support dialogue formalizes the task and introduces benchmark datasets grounded in psychological theories, such as \textsc{ESConv} based on Helping Skill theory~\cite{esc}. Building on this foundation, subsequent studies expand the strategy space through synthetic data generation, exemplified by \textsc{ExTES}~\cite{extes}. To better select support strategies, recent work adopts reasoning mechanism to analyze the strategies embedded in emotional support conversations~\cite{zhao2023transesc, escot, zhao2025chain}. Additionally, \citet{sweet} improve diversity and realism by employing multi-agent role-playing to simulate complex support scenarios, while \citet{bhdt} further consider the trade-off between support effectiveness and user effort. However, existing studies primarily focus on single-session interactions, leaving the problem of building long-term supportive relationships with users largely unexplored. This gap motivates the need for a theoretically grounded benchmark to assess how agents leverage memory to provide emotionally appropriate support in long-term scenarios.

\section{ENPMR-Bench Overview}

\subsection{Task Formulation}

Each instance in ENPMR-Bench is equipped with a detailed user profile, which contains basic demographic information (such as name, age, occupation) as well as a set of user-related memory entries. To systematically evaluate large language models’ capability to autonomously analyze user needs and determine which memories to retrieve in emotional support scenarios, we design a structured memory recall framework and a series of short-turn emotional support dialogues that integrate memory.

Specifically, in each round of the emotional support dialogue, we explicitly require the agent to recall a relevant memory. The agent is provided with the dialogue history, user profile, and available memory entries, and is tasked with selecting an appropriate memory and generating a supportive response that incorporates this memory. This process is formally defined as follows:

\begin{equation}
    a_{t+1}, m = \text{LLM}(H, P, M)
\end{equation} where, given the static user profile $P$, the current dialogue history $H = \{u_1, a_1, \dots, u_t\}$, and the memory bank $M = \{m_1, m_2, \dots, m_N\}$, the agent is expected to select a memory item $m$ and generate the next utterance $a_{t+1}$.

Our benchmark primarily focuses on assessing whether the model can accurately identify the underlying user needs that trigger negative emotions, recall the corresponding type of memory based on these needs, and employ the recalled memory to deliver more effective emotional support.

\subsection{Structured Memory Retrieval Guidelines}

To address the high cost and annotation difficulty of collecting emotional support dialogue data, we ground our design in Maslow’s hierarchy of needs~\cite{maslow} and collaborate with human experts to develop a structured memory retrieval framework. This framework guides both the construction of emotional support scenarios and the generation of memory-augmented dialogues with large language models. As shown in Table~\ref{tab:need_memory_recall}, the framework organizes memory retrieval into four core need dimensions, namely physiological needs, love and belonging, esteem, and self-actualization. Each dimension is associated with characteristic user behaviors under unmet conditions and corresponding memory types. The design process was conducted in close collaboration with human experts to ensure theoretical soundness. Importantly, distinct unmet needs are associated with different patterns of memory recall.

\begin{figure*}[ht!]\centering
    \includegraphics[width=1.0\textwidth]{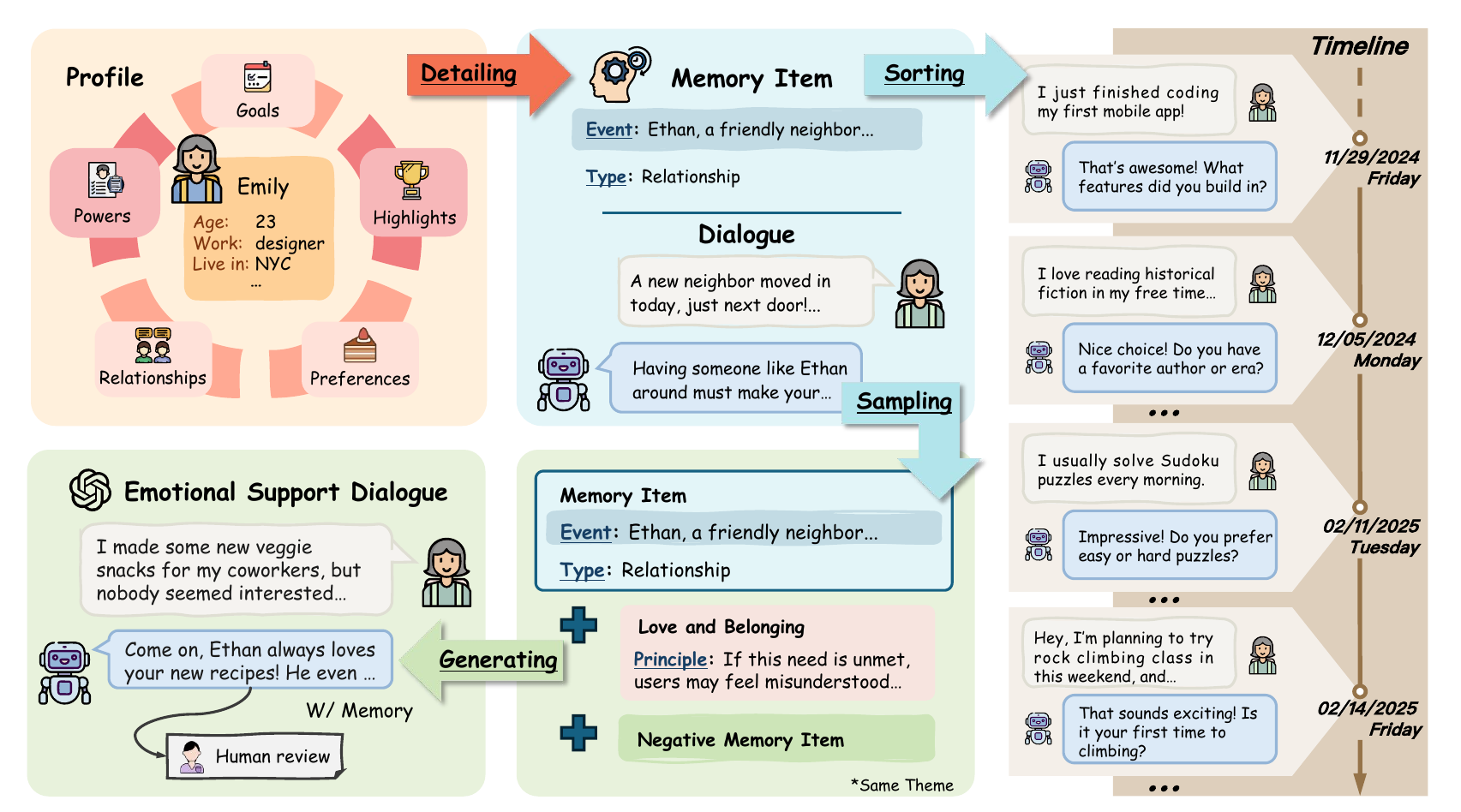}
    \caption{The data construction process of \textsc{ENPMR-Bench} involves the construction of a structured, multi-dimensional user profile organized around central life themes. Memories are chronologically arranged to form a temporal narrative. Subsequently, relevant memories are sampled to design emotional support scenarios, which are then used to generate appropriate dialogues.}
    \label{fig:main}
\end{figure*}

\subsection{Data Statistics}

ENPMR-Bench, a Chinese benchmark, consists of 11,846 memory entries across five predefined categories—highlight (3,729 entries), power (3,463), relationship (2,324), goal (1,719), and preference (611). Each entry is paired with a dialogue history comprising 10 to 20 turns between the user and the AI, covering ten life themes tailored to the user's profile. The benchmark also includes 1,872 emotional support dialogues, averaging four turns per dialogue. These dialogues correspond to four levels of user needs, including Self-Actualization (1,099 instances), Esteem (401), Physiological (192), and Love and Belonging (180). Each dialogue is grounded in a single user memory.

\section{Data Construction}

\subsection{User Profile Generation}

We build upon the persona repository from PersonaHub~\cite{personahub}, randomly sampling a set of initial personas characterized by multiple attributes, including gender, age, occupation, interests, and lifestyle. Each persona is described using natural language and then converted into a structured user profile. To enrich these profiles, we automatically generate ten personalized life themes for each persona, reflecting major events and salient concerns throughout the user's life course. Each theme corresponds to specific goals, challenges, or values (e.g., participation in community events, pursuit of musical achievements), ensuring mutual independence while collectively covering diverse aspects of the persona's needs. For each life theme, we systematically generate and annotate key memory entries with temporal information, encompassing the following categories:

\begin{itemize}
    \item \textbf{Relationships}: Covering both positive and strained social ties as well as representative events, reflecting the user's social support network, conflicts, and interaction patterns.
    \item \textbf{Powers}: Capturing the persona's strengths and weaknesses within the given life theme, mapping both the skills they excel at and those they struggle with.
    \item \textbf{Goals}: Identifying the goals pursued by the user within the given life theme, including both those that have been achieved and those that remain unfulfilled, thereby reflecting their evolving aspirations and priorities.
    \item \textbf{Highlights}: Documenting pivotal moments when the user received external recognition or achieved personal breakthroughs, showcasing their positive accomplishments.
\end{itemize}

In addition, we maintain a separate category of memory entries, \textbf{Preferences}, which record user-specific likes and dislikes (e.g., dietary habits, hobbies, and other personal inclinations). Unlike other memory categories that are associated with specific life themes, preferences capture user tendencies that influence behavior and decision-making.

\begin{itemize}
    \item \textbf{Preferences}, which stores user-specific likes and dislikes, such as dietary preferences, hobbies, and other personalized inclinations.
\end{itemize}

The structured user profiles form the basis for test case generation, enabling the design of highly targeted reference answers anchored in persona-specific experiences and traits.

\subsection{Generate Sessions From Profile}

To convert enriched user profiles into interactive session data, we systematically transform each memory entry into user–AI dialogue snippets that simulate realistic and evolving conversations. Each memory entry serves as a narrative anchor: the user's utterances naturally reference or reveal the underlying event, goal, or experience, while the AI responds with context-aware and persona-aligned replies. These dialogue snippets are then concatenated in strict temporal order according to the timestamps of the memory entries, thereby reconstructing a coherent conversational history that unfolds along the persona's life trajectory.

\subsection{Emotional Support Dialogue Generation}

We construct emotional support dialogues grounded in the most recent memory entry of the user profile, simulating future support scenarios where the agent has full access to user history. Following our structured memory retrieval protocol, we randomly sample a target memory entry according to the user’s pre-specified need. To enhance specificity, we further employ negative sampling by selecting additional memory entries that share similar life themes but differ in concrete details as negative samples for each target memory. These negative samples serve as distractors, encouraging GPT-4o to generate dialogues that are explicitly grounded in the target memory rather than generic to the theme. The model is prompted to generate a concise dialogue turn that references only the target memory, explicitly avoiding information from negative samples.

\subsection{Data Filtering and Human Validation}

We employ GPT-4o to automatically evaluate the quality of synthesized dialogues using the BLRI scale~\citep{barrett1962dimensions}, filtering out instances with low scores. The filtered data is then subjected to human validation to ensure consistency and reliability. Additionally, we randomly sample dialogues for human evaluation, where two annotators independently rate each response (0 or 1) based on three criteria: appropriateness, relevance, and correctness. The results demonstrate robust model performance, with average scores of 0.80 (90\% agreement), 0.93 (95\% agreement), and 0.93 (85\% agreement), respectively. Further details are provided in the Appendix~\ref{sec: human_evaluation}.

\begin{table*}[ht!]
    \centering
    \small
    \renewcommand{\arraystretch}{1.15}
    
    \begin{tabularx}{\textwidth}{c|YYYY|YYYY|YYYY}
    \toprule
    \multirow{2}{*}{\textbf{Embedding Models}} & \multicolumn{4}{c|}{\textbf{Recall}} & \multicolumn{4}{c|}{\textbf{Precision}} & \multicolumn{4}{c}{\textbf{nDCG}} \\
    & @1 & @3 & @5 & @10 & @1 & @3 & @5 & @10 & @1 & @3 & @5 & @10 \\
    \midrule
    gte-Qwen2-7B-instruct       & 6.30 & 17.95 & 26.82 & 40.87 & 6.30 & 5.98 & 5.36 & 4.09 & 6.30 & 12.97 & 16.60 & 21.15 \\
    Qwen3-Embedding-8B          & 9.46 & 22.97 & 30.77 & 46.42 & 9.46 & 7.66 & 6.15 & 4.64 & 9.46 & 17.22 & 20.41 & 25.44 \\
    doubao-embedding & 9.29 & 21.85 & 30.50 & 44.82 & 9.29 & 7.28 & 6.10 & 4.48 & 9.29 & 16.47 & 20.00 & 24.62 \\
    text-embedding-3-large      & 7.05 & 19.12 & 26.28 & 40.76 & 7.05 & 6.37 & 5.26 & 4.08 & 7.05 & 14.04 & 17.00 & 21.67 \\
    \bottomrule
    \end{tabularx}%
    
    \caption{Retrieval performance comparison of different embedding models on ENPMR-Bench.}
    \label{tab:embedding_model_performance}
\end{table*}

\begin{table*}[ht!]
    \centering
    \small
    \renewcommand{\arraystretch}{1.15}
    \begin{tabularx}{\textwidth}{cc|c|YYYYY|YYYYY}
    \toprule
    \multirow{2}{*}{\textbf{LLMs}} & \multirow{2}{*}{ } & \multirow{2}{*}{ } & \multicolumn{5}{c|}{\textbf{GPT}} & \multicolumn{5}{c}{\textbf{Human}} \\
    & & TMR & Flu. & Hum. & Info. & Emp. & Avg. & Flu. & Hum. & Info. & Emp. & Avg. \\
    \midrule
    \multirow{3}{*}{Qwen-Max}
    & base      
    & 0.49 & 4.61 & 4.23 & 3.41 & 4.16 & 4.10 
    & 4.52 & 4.18 & 3.45 & 4.03 & 4.05 \\
    & w/CoT     
    & 0.58 & 4.67 & 4.26 & 3.50 & 4.24 & 4.17 
    & 4.50 & 4.17 & 3.35 & 4.08 & 4.03 \\
    & w/Golden  
    & -- & 4.96 & 4.75 & 4.13 & 4.88 & 4.68 
    & 4.93 & 4.72 & 4.17 & 4.68 & 4.62 \\
    \midrule

    \multirow{3}{*}{DeepSeek-V3}
    & base      
    & 0.53 & 4.77 & 4.89 & 3.93 & 4.59 & 4.55 
    & 4.43 & 4.47 & 3.75 & 4.23 & 4.22 \\
    & w/CoT     
    & 0.59 & 4.81 & 4.90 & 3.98 & 4.66 & 4.59 
    & 4.52 & 4.68 & 3.97 & 4.42 & 4.40 \\
    & w/Golden  
    & -- & 4.95 & 4.98 & 4.36 & 4.91 & 4.80 
    & 4.75 & 4.83 & 4.35 & 4.63 & 4.64 \\
    \midrule

    \multirow{3}{*}{GPT-4o}
    & base      
    & 0.52 & 4.57 & 4.33 & 3.44 & 4.16 & 4.12 
    & 4.35 & 4.10 & 3.37 & 3.93 & 3.94 \\
    & w/CoT     
    & 0.59 & 4.64 & 4.37 & 3.51 & 4.23 & 4.19 
    & 4.53 & 4.27 & 3.53 & 4.20 & 4.13 \\
    & w/Golden  
    & -- & 4.85 & 4.64 & 3.85 & 4.68 & 4.51 
    & 4.77 & 4.63 & 3.95 & 4.47 & 4.45 \\
    \midrule
    
    \multirow{3}{*}{Gemini-2.5-Flash}
    & base      
    & 0.50 & 4.60 & 4.54 & 3.63 & 4.34 & 4.28 
    & 4.53 & 4.55 & 3.53 & 4.12 & 4.18 \\
    & w/CoT     
    & 0.59 & 4.65 & 4.67 & 3.69 & 4.42 & 4.36 
    & 4.50 & 4.45 & 3.50 & 4.23 & 4.17 \\
    & w/Golden  
    & -- & 4.85 & 4.78 & 4.05 & 4.77 & 4.61 
    & 4.87 & 4.85 & 4.02 & 4.58 & 4.58 \\
    
    \bottomrule
    \end{tabularx}%
    
    \caption{Performance comparison of different LLMs on ENPMR-Bench. Responses are evaluated along four core dimensions by both GPT-4o and human annotators. Detailed descriptions of the evaluation prompts and analyses of the agreement between human and GPT are provided in the Appendix~\ref{sec:human_eval}}
    \label{tab:llms_performance}
\end{table*}

\section{Experiment}
\subsection{Memory Retrieval Task}

\textbf{Task Definition and Models.}
This task evaluates a model’s ability to retrieve the most relevant past interaction from a user’s memory given an ongoing dialogue. Let $H=\{u_1,a_1,\dots,u_t\}$ denote the dialogue history and $M=\{m_1,\dots,m_N\}$ the memory bank, where each memory entry is a multi-turn dialogue snippet. The objective is to select the memory $m^*\in M$ that best matches the current context. We evaluate several strong Chinese embedding models, including qwen3-embedding-8B~\cite{qwen3embedding}, gte-Qwen2-7B-instruct, doubao-embedding, and text-embedding-3-large. 

\textbf{Metrics.}
To comprehensively evaluate the performance of the embedding model in memory retrieval tasks, we computed multiple metrics, including nDCG, Recall, and Precision. nDCG accounts for both relevance and ranking position. 

\subsection{Response Generation Task}

\textbf{Task Definition and Models.}
This task assesses a model’s ability to generate contextually appropriate and personalized emotional support responses grounded in user profiles and retrieved memories. 

For the LLM-based evaluation setting, we select a range of representative models, including GPT-4o, Deepseek-v3~\citep{liu2024deepseek}, Qwen-Max, and Gemini-2.5-Flash. For each model, we consider three evaluation settings: (i) base, where the model autonomously selects a single memory for response generation; (ii) w/CoT, where the model is provided with explicit step-by-step reasoning instructions to guide memory selection; and (iii) w/Golden, where the model is directly given the golden memory annotated in ENPMR-Bench.

For the Agentic Memory evaluation, we adopt the current state-of-the-art systems, including Mem0~\citep{chhikara2025mem0}, Memu~\citep{memu2025}, and MemOS~\citep{li2025memos}. Consistent with prior work, all these memory systems utilize GPT-4o-mini to construct their memory banks. In both the RAG and Agentic Memory settings, Deepseek-v3 is employed as the response generation model, due to its demonstrated strength in Chinese empathetic dialogue. All generations use a temperature of 0.8.

\noindent \textbf{Metrics.}
Following the LLM-as-judge paradigm, responses are assessed along four dimensions, including Fluency, Humanoid, Information and Empathy~\cite{esceval}, with empathy criteria adapted from established psychological scales~\cite{clark1998impact}. To ensure stable scoring, we set the temperature of GPT-4o to 0 during the evaluation process. In addition, we introduce the "Memory Type Match Rate" (TMR.) to evaluate whether the memory utilized by the model during response generation conforms to the predefined memory retrieval principles, specifically whether the selected memory type aligns with the expected memory category. Since agentic memory mechanisms merge and reorganize historical information, this metric is applied only in the LLM-based evaluation setting.

In practice, LLMs typically select memories from a pre-filtered set retrieved by an external memory module. Accordingly, LLM-based evaluation is conducted over a memory subset sharing the same topic as the current dialogue history. In contrast, evaluations under the RAG and Agentic Memory settings are conducted over the full memory bank. Details are provided in the Appendix~\ref{sec:llm-as-judge}.

\section{Result and Analysis}

To systematically analyze where current systems succeed and fail in proactive memory retrieval for emotional support, we decompose the problem into four sequential research questions:

RQ1: How do existing memory systems perform in retrieving relevant memories for emotional support scenarios?

\begin{figure*}[ht!]
    \centering
    \subfigure[GPT-4o]{
        \includegraphics[width=0.23\linewidth]{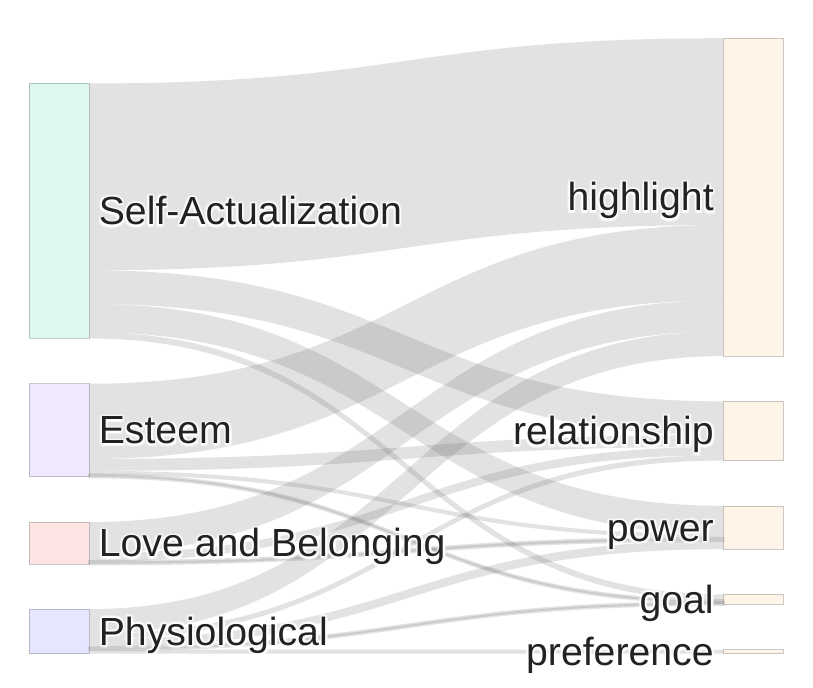}
    }
    \subfigure[Gemini-2.5-Flash]{
        \includegraphics[width=0.23\linewidth]{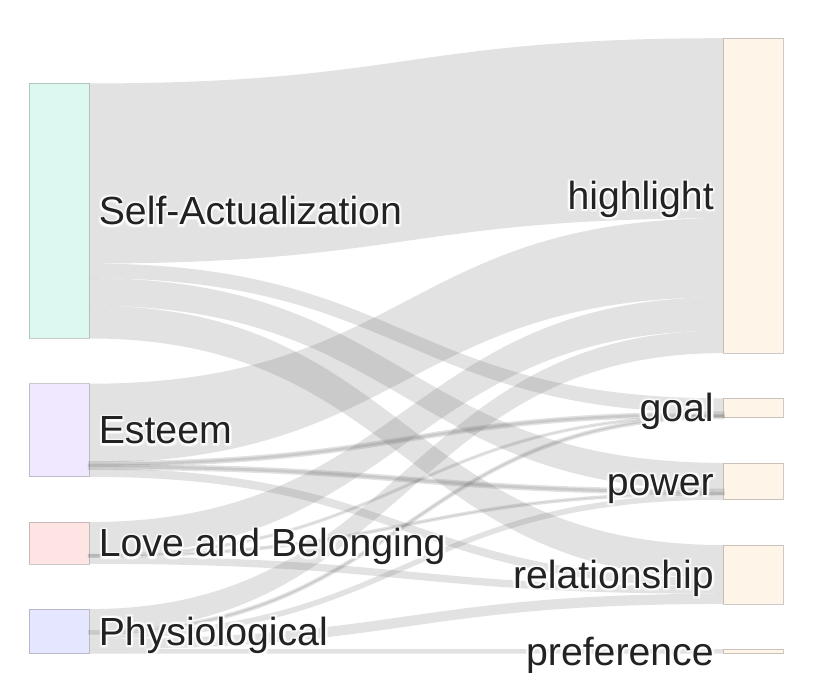}
    }
    \subfigure[Qwen-Max]{
        \includegraphics[width=0.23\linewidth]{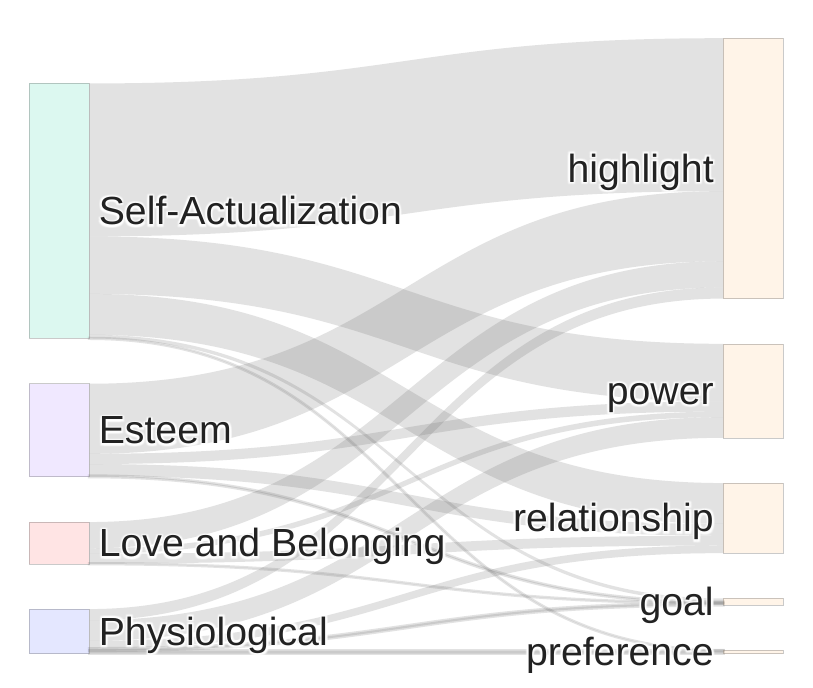}
    }
    \subfigure[Deepseek-V3]{
        \includegraphics[width=0.23\linewidth]{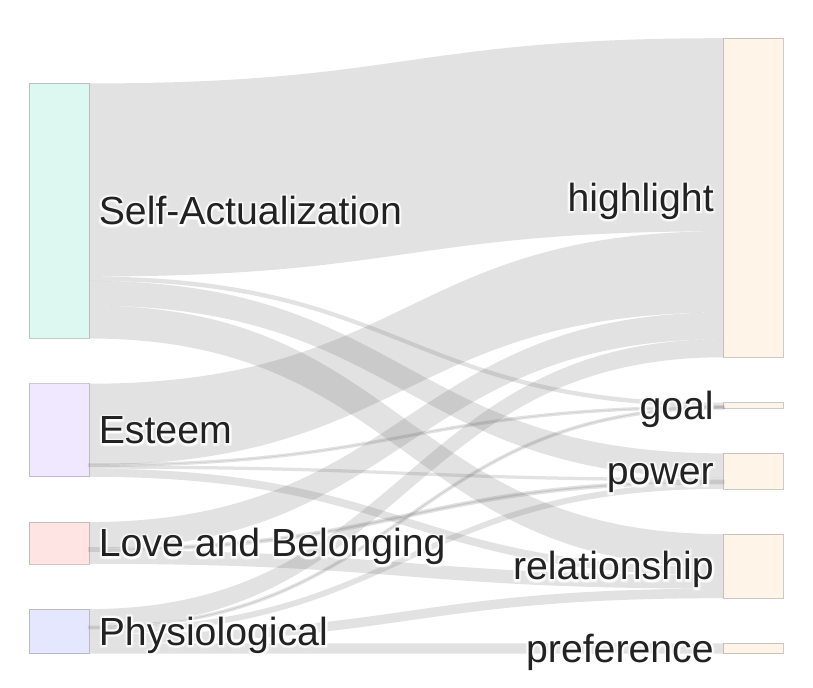}
    }
    \\
    \subfigure[gte-Qwen2-7B-instruct]{
        \includegraphics[width=0.23\linewidth]{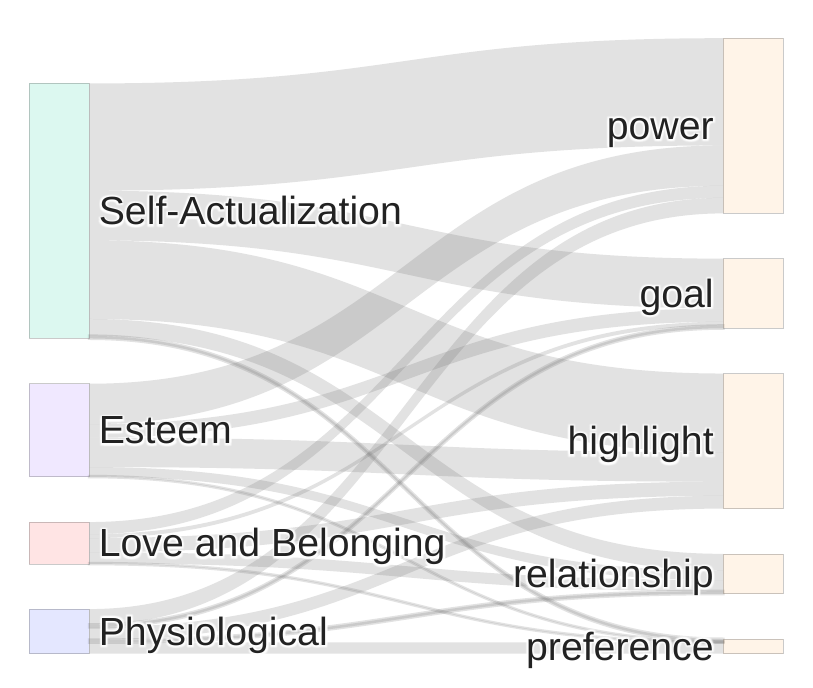}
    }
    \subfigure[Qwen3-Embedding-8B]{
        \includegraphics[width=0.23\linewidth]{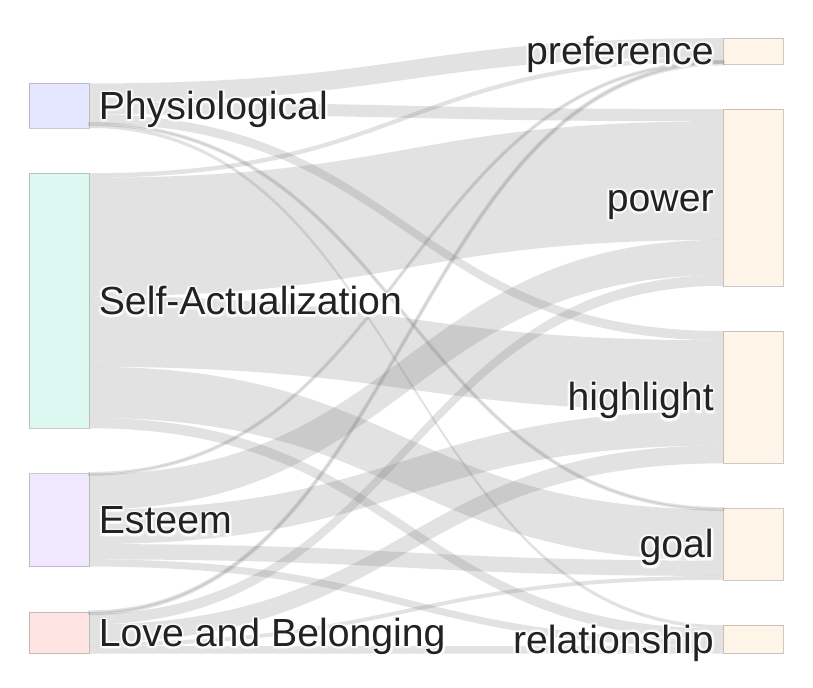}
    }
    \subfigure[doubao-embedding]{
        \includegraphics[width=0.23\linewidth]{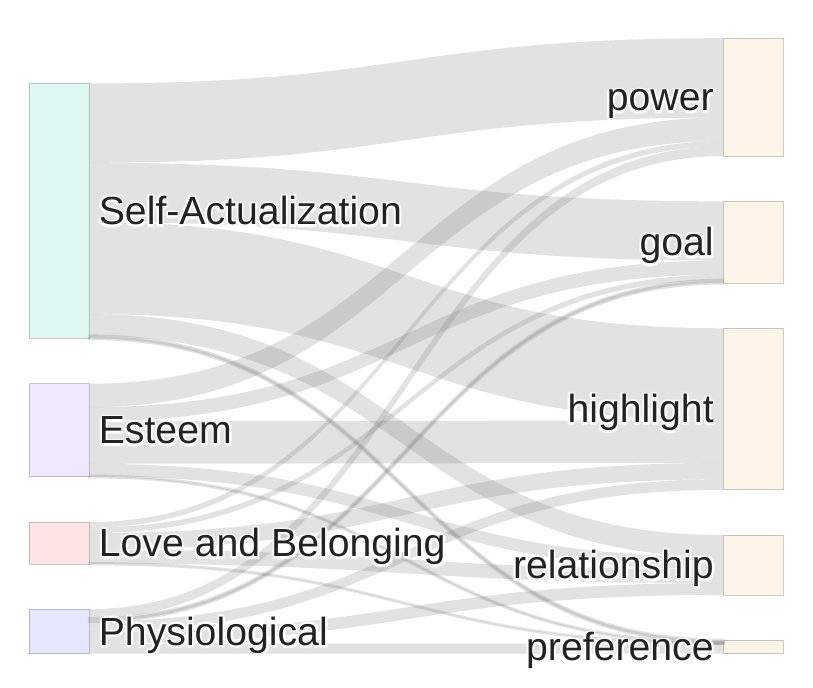}
    }
    \subfigure[text-embedding-3-large]{
        \includegraphics[width=0.23\linewidth]{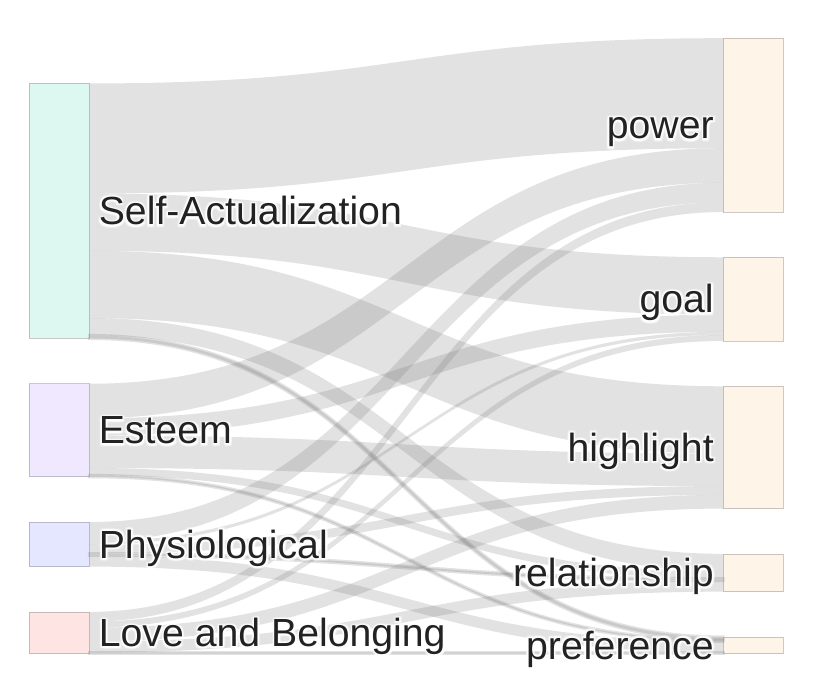}
    }
    \caption{Sankey diagrams illustrating the mapping between human user needs (left) and the most frequently retrieved memory types (Top-1, right) across four LLMs and embedding models. Flow width indicates retrieval frequency, enabling direct comparison of how each model aligns user needs with memory types.}
    \label{fig:sankey_llm}
\end{figure*}

\begin{table}[ht!]
    \centering
    \small
    \renewcommand{\arraystretch}{1.15}
    \begin{tabularx}{\linewidth}{c|YYYYY}
    \toprule
    \multirow{2}{*}{\textbf{}} & \multicolumn{5}{c}{\textbf{GPT}} \\
    & Flu. & Hum. & Info. & Emp. & Avg. \\
    \midrule
    \rowcolor{gray!20}
    \multicolumn{6}{c}{\rule{0pt}{2.4ex}\textbf{RAG}} \\
    gte-Qwen2-7B-instruct   & 4.60 & 4.74 & 3.99 & 4.33 & 4.41 \\
    Qwen3-Embedding-8B      & 4.76 & 4.76 & 4.02 & 4.36 & 4.43 \\
    doubao-embedding        & 4.63 & 4.75 & 4.03 & 4.38 & 4.45 \\
    text-embedding-3-large  & 4.59 & 4.75 & 3.95 & 4.36 & 4.41 \\
    \midrule
    \rowcolor{gray!20}
    \multicolumn{6}{c}{\rule{0pt}{2.4ex}\textbf{Agentic Memory}} \\
    Mem0       & 4.00 & 4.48 & 3.27 & 3.96 & 3.93 \\
    MemOS      & 4.65 & 4.78 & 3.95 & 4.48 & 4.46 \\
    Memu       & 4.55 & 4.74 & 3.86 & 4.35 & 4.37 \\
    
    \bottomrule
    \end{tabularx}%

    \caption{Performance of generated responses by different memory methods on ENPMR-Bench, using DeepSeek-V3 as the generation model.}
    \label{tab:am_performance}
\end{table}

RQ2: To what extent does retrieval quality influence the emotional effectiveness of generated responses?

RQ3: To what extent can current LLMs effectively utilize retrieved memories to generate supportive responses?

RQ4: Does ENPMR-Bench provide a valuable and challenging resource for evaluating memory-augmented agents' emotional intelligence?

\subsection{RQ1: Limited Effectiveness of Emotional Need-Aware Memory Retrieval}

\textbf{Semantic relevance alone is insufficient for emotion-grounded memory retrieval.}
Table~\ref{tab:embedding_model_performance} shows that embedding-based retrieval performs poorly on ENPMR-Bench. Even the best model reaches only 46.41\% Recall@10, while Top-1 accuracy remains below 10\% across all models. These results indicate a limitation of semantic similarity in identifying emotionally salient memories that are critical for grounding empathetic responses.

\begin{table}[ht!]
\centering
\small
\renewcommand{\arraystretch}{1.15}
\resizebox{\linewidth}{!}{
\begin{tabular}{ccccc}
\hline
\textbf{} & \textbf{PN} & \textbf{LB} & \textbf{EN} & \textbf{SA} \\
\hline
Qwen-Max          & 4.22 & 3.99\Lo*** & 4.57\Hi*** & 4.17 \\
DeepSeek-V3       & 4.53 & 4.34\Lo* & 4.85\Hi*** & 4.66 \\
GPT-4o            & 4.13 & 3.82\Lo*** & 4.56\Hi*** & 4.19 \\
Gemini-2.5-Flash  & 4.14 & 4.01\Lo* & 4.73\Hi*** & 4.43 \\
\hline
\end{tabular}
}
\caption{Empathy scores of LLMs across different emotional need dimensions. \emph{L}/\emph{H} mark the lowest/highest average score of each model. Asterisks denote statistically significant differences between an extreme score and its adjacent score in rank within the same model(*: $p<0.05$, **: $p<0.01$, ***: $p<0.001$).}
\label{tab:need_empathy_scores}
\end{table}

\textbf{High-level memory aggregation yields limited gains.}
Agentic memory systems moderately improve empathy by aggregating historical interactions at a higher level. As shown in Table~\ref{tab:am_performance}, MemOS achieves an empathy score of 4.48, surpassing the best RAG-based result (4.38). However, a clear gap remains compared to the w/Golden setting in Table~\ref{tab:llms_performance}, suggesting that current agentic memory systems still fall short in emotionally adaptive memory retrieval.

\begin{table*}[h]
    \centering
    \small
    \renewcommand{\arraystretch}{1.15}
    \begin{tabularx}{\textwidth}{@{} >{\bfseries}c X @{}}
    \toprule
    \multicolumn{2}{{@{} p{\linewidth} @{}}}{\textbf{Scene:} After a local charity photography event wrapped up, Lin Xiao opened up to the AI about feeling misunderstood and isolated. Some audience members questioned his environmental theme, which left him craving genuine understanding and acceptance of his artistic expression. Being in an unfamiliar community only deepened his sense of not belonging.} \\
    \multicolumn{2}{{@{} p{\linewidth} @{}}}{\textbf{History:} \textit{User: The charity photography event ended today. Some people were really cold about my environmental theme and even said art has nothing to do with environmentalism. Honestly, I’m pretty down. It feels like no one really gets what I’m trying to express. }} \\
    \midrule
    \midrule
    \multicolumn{1}{c}{\textbf{Model}} & \multicolumn{1}{c}{\textbf{Model Response}} \\
    \midrule
    Base & Oh come on, Lin Xiao—how could that be? Hearing you say that honestly breaks my heart. You’re the designer who can turn scrap materials into masterpieces—how could no one get the deeper meaning behind your work? Don’t be discouraged. You’re not fighting this battle alone. \\
    \hline
    w/CoT & Lin Xiao, don’t be too down. \textbf{Remember that little girl} at your Green Vision exhibition—the one who asked her mom, ‘Can we use fewer plastic bags?’ Even the smallest seed can take root and grow in someone’s heart. Your photography is like planting flowers in the cracks of concrete. Even if some people don’t get it right now, that life force will find its own way to grow. \\
    \hline
    w/Golden & Lin Xiao, don’t be discouraged. \textbf{Professor Zhang} has always believed that design should talk to society, and he really appreciates how you mix environmental ideas with visual language. He’d definitely get how you’re feeling. \textbf{Maybe talk to him about it}, it could give you some fresh inspiration and motivation. \\
    \bottomrule
    \end{tabularx}
    \caption{A representative example of an output generated by Gemini-2.5-Flash. In this case, the user’s underlying need is love and belonging. The base model fails to recognize this emotional need, whereas the w/CoT variant successfully infers it and retrieves a semantically and emotionally aligned memory to support the response.}
    \label{tab:case_study}
\end{table*}

\textbf{LLMs exhibit systematic biases in memory selection.}
As shown in Table~\ref{tab:llms_performance}, even with a highly constrained and relevant retrieval space, LLMs achieve a maximum TMR of only 0.53, indicating limited need-oriented memory selection even within a constrained retrieval space. CoT guidance yields consistent but modest gains, yet a substantial gap remains to the w/Golden upper bound. In addition, as illustrated in Figure~\ref{fig:sankey_llm}, sankey analyses of Top-1 retrieval further reveal model-specific preferences that deviate from need-appropriate patterns. Particularly, for love and belonging needs, models systematically over-select highlight memories rather than the memories of relationship. This misalignment is further reflected in Table~\ref{tab:need_empathy_scores}, where love and belonging consistently receive the lowest empathy scores across models. This evidence suggests that systematic selection preferences interact unevenly with different emotional needs, leading to imbalanced empathetic performance.

\subsection{RQ2\&4: Appropriate Memory Is Critical for Empathetic Response}

\textbf{Emotionally mismatched memories undermine empathetic response quality.} As shown in Table~\ref{tab:llms_performance}, although current models achieve strong fluency and humanoidness, grounding responses in inappropriate memories consistently leads to lower empathy and coherence scores. Across different configurations of DeepSeek-V3, both RAG-based and agentic memory settings remain substantially inferior to the w/Golden upper bound, indicating that retrieval quality is the primary bottleneck. Incorporating chain-of-thought reasoning yields consistent but limited improvements by better aligning retrieved memory types with inferred user needs. These findings underscore that memory appropriateness, rather than surface-level generation capabilities, is crucial for producing emotionally grounded responses. Moreover, these results also highlight the benchmark's ability to expose subtle deficiencies in memory utilization within existings agents. 

\subsection{RQ3: Limitations of LLMs in Leveraging Retrieved Memories}

Notably, even under the w/Golden setting, in which memory selection is fully controlled, models exhibit non-trivial gaps in empathy scores across multiple evaluation dimensions. This suggests that limitations persist at the generation stage, particularly in effectively integrating memory content into emotionally grounded and contextually appropriate responses. Similar trends are also consistently observed in human evaluation results. 

\section{Conclusion}

In this work, we introduce ENPMR-Bench, a novel benchmark designed to evaluate the agents' capacity of ENPMR. Through extensive experiments, we show that, despite recent advances in memory retrieval and response generation, current agents remain inadequate in emotionally complex scenarios that require affective reasoning. In particular, agents frequently fail to align user emotional needs with appropriate memory types. Further analysis reveals that these shortcomings are not random but stem from systematic biases in memory selection. We hope that our benchmark opens new avenues for future exploration and advances the development of more empathetic, memory-augmented agents.

\section*{Limitations}

This work focuses on evaluating an agent’s ability to proactively infer users’ latent emotional needs and retrieve need-appropriate memories to support emotionally grounded dialogue. Rather than modeling naturalistic emotional conversations in the wild, ENPMR-Bench is designed as a diagnostic benchmark that isolates and examines the alignment between emotional need inference and memory retrieval under theoretically grounded constraints. Within this scope, several limitations remain:

\begin{itemize}
    \setlength{\itemsep}{2pt}
    \item \textbf{Synthetic Data Construction:}
    ENPMR-Bench is primarily constructed using LLMs, with human involvement limited to expert-guided design and small-scale validation. In addition, the empathetic dialogues are relatively short compared to real-world emotional support interactions, which often involve longer and more iterative exchanges.

    \item \textbf{Theory-Grounded Memory Mapping:}
    The mapping between emotional needs and memory types, as well as the TMR metric, is derived from psychological theory and expert-informed design rather than empirical optimality. As a result, it may not fully capture the diversity of memory usage patterns in real-world emotional support scenarios.

\end{itemize}

Future work may incorporate broader human supervision and longer multi-turn interactions to better capture dynamic emotional needs and more rigorously validate need–memory alignment in real-world settings.

\section*{Ethical Statement}

We are committed to publicly releasing all data upon acceptance of the paper. All experts participating in this study hold at least a bachelor’s degree and possess a minimum of two years of relevant professional experience. In particular, the experts who participated in the formulation of the memory retrieval principles possess substantial experience in emotional support–related practice. Their compensation is calculated based on the hours worked and is aligned with the average income levels for similar professions in the region. Furthermore, we are fully aware of the potential biases associated with LLM-as-Judge. To mitigate these effects, we incorporated human expert assessments. However, due to cost considerations, the scale of human evaluation remains limited at this stage. We note that this constraint is common in current conversational AI research that relies on LLMs.

\section*{Acknowledgements}

We thank the anonymous reviewers for their comments and suggestions. This work was supported by the National Natural Science Foundation of China (NSFC) via grant 62441614 and 62576125.

\bibliography{acl}

\appendix

\section{Data Construction}
\subsection{Dialogue Generation}

\textbf{Dialogue Generated by LLM.} Following prior work on LLM-based construction of emotional support dialogues, we adopt a script-based generation framework to ensure controllability and internal coherence. Specifically, conditioned on a specified emotional need, the LLM is first prompted to construct a complete interaction scenario, including both the situational context and the user's emotional state. The model then performs a self-consistency check to assess whether the generated scenario is plausible and internally coherent. Based on the validated scenario, the dialogue is subsequently generated, conditioned on the predefined emotional need and the associated memory grounding. LLMs exhibit strong capabilities in producing natural, fluent, and contextually appropriate dialogues when guided by explicit design principles and constraints. And, we recognize the potential limitations and biases introduced by LLM-generated data, particularly in affective domains. To mitigate these risks, we incorporate human validation at critical stages of data construction, serving as a quality control mechanism. The prompts used for synthetic data generation are shown in Figures~\ref{fig:prompt1}-~\ref{fig:prompt6}.

\noindent \textbf{Negative Sampling.} To prevent models from generating responses that are only loosely aligned with the target memory at a thematic level, we adopt a negative sampling strategy during emotional support dialogue construction. For each target memory, we further incorporate two to three negative memory entries with the same life theme and memory category but differ in specific events or details.

\subsection{Data Filtering}

To ensure the faithfulness and emotional validity of the constructed dialogues, we adopt a two-stage data filtering procedure. In the first stage, we apply a GPT-based self-check mechanism to verify whether each generated dialogue is faithful to the associated memory item. In the second stage, we perform scale-based quality filtering using a BLRI Scale. Each remaining dialogue is scored across multiple dimensions related to emotional support quality, and instances with an average score lower than 2 are discarded. Through this two-stage filtering process, we retain only dialogues that are both memory-grounded and emotionally meaningful, thereby ensuring that downstream evaluation reflects models' true capabilities in need-aware memory retrieval.

\begin{figure}[h!]
\begin{prompt}{BLRI Scale}
[Question List]
1. The counselor usually senses or is aware of my feelings at the moment.\\
2. The counselor reacts to my words, but does not understand how I feel inside.\\
3. The counselor almost always understands exactly what I mean.\\
4. The counselor can accurately empathize with the feelings that my experiences arouse in me.\\
5. The counselor does not understand me.\\
6. The counselor's own attitudes towards certain things get in the way of his/her understanding me.\\
7. The counselor is aware of what I mean even when I have difficulty expressing it.\\
8. The counselor does not listen to me and does not grasp my thoughts and feelings.\\
9. The counselor usually understands the whole of what I mean.\\
10. The counselor is not aware of how sensitive I am to some of the things we discuss.\\
11. The counselor's responses to me are so fixed and automatic that I feel I cannot really communicate with him/her.\\
12. When I feel hurt or upset, the counselor can accurately identify my painful feelings without becoming upset himself/herself.\\
\end{prompt}
\caption{Prompt for Data Filtering.}\label{fig:prompt_df}
\end{figure}

\subsection{Human Validation Protocol}
\label{sec: human_evaluation}

\begin{figure}[h]
    \centering
    \includegraphics[width=\linewidth]{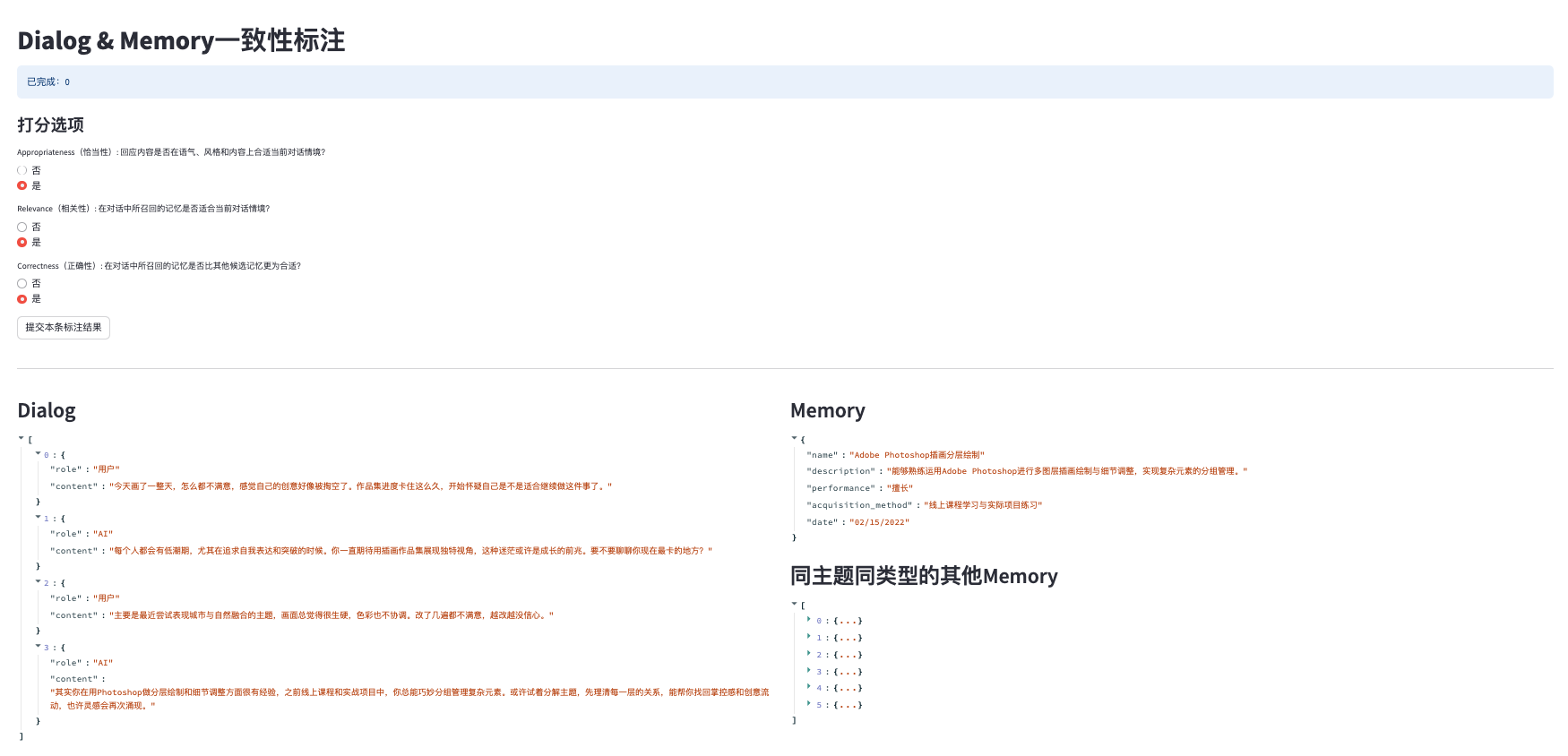}
    \caption{The interface used by human annotators}
    \label{fig:human_evaluation}
\end{figure}

To assess the overall quality of the generation process, we randomly sampled a subset of generated entries and conducted a small-scale human evaluation with recruited volunteers. We developed a visual evaluation interface using Streamlit, as shown in Figure~\ref{fig:human_evaluation}. For each entry, the evaluators were asked to perform binary (yes/no) assessments on the following three dimensions:

\begin{itemize}
    \setlength{\itemsep}{1pt}
    \item \textbf{Appropriateness}: Whether the response is suitable in tone, style, and content for the given dialogue context.
    \item \textbf{Relevance}: Whether the retrieved memory aligns with the current dialogue context.
    \item \textbf{Correctness}: Whether the retrieved memory is more appropriate than the candidate memories.
\end{itemize}

Due to the challenges of reading long texts, we did not require the evaluators to select the most appropriate memory entry. Instead, we directly provided the dialogue and the corresponding metadata to the evaluators.

\section{Evaluation Setting}

\subsection{LLM-as-Judge Setting}
\label{sec:llm-as-judge}
In the LLM-as-Judge setting, we adopt GPT-4o as an automatic evaluator to assess the quality of generated emotional support responses. Following prior work, each response is evaluated along four dimensions: Fluency, Humanoid, Information, and Empathy. \textbf{Fluency} evaluates the smoothness and naturalness of expression in dialogues.  \textbf{Humanoid} examines the distinction between emotional assistants and human conversational behavior. \textbf{Empathy} assesses the system’s ability to comprehend user emotions and accurately capture the underlying emotional logic. \textbf{Information} evaluates the relevance and adequacy of recommendations provided by the emotional assistant. The specific prompt is shown in Figure~\ref{fig:prompt10}.

\subsection{Human Evaluation}
\label{sec:human_eval}
\begin{table}[h]
    \centering
    \resizebox{\linewidth}{!}{
    \begin{tabular}{lcc}
        \toprule
        & \textbf{GPT vs. Annotator 1} & \textbf{GPT vs. Annotator 2} \\
        \midrule
        $\kappa$ & 0.765 & 0.748 \\
        \bottomrule
    \end{tabular}
    }
    \caption{Inter-annotator agreement measured by Cohen’s kappa between GPT and human annotators.}
    \label{tab:inter_annotator_agreement}
\end{table}

Human evaluation is conducted to validate the credibility of automatic scoring and to examine consistency between model-based and human judgments. Two annotators independently assess a randomly sampled subset of responses following the same evaluation criteria. Agreement between GPT-based scores and each human annotator is measured using quadratic weighted Cohen’s kappa, with results indicating substantial agreement across annotators. These findings suggest that GPT-based evaluation does not systematically deviate from human perception when assessing emotionally grounded responses, while human evaluation serves as an essential reference for interpreting LLM-based scores.

\subsection{Significance Testing}

To investigate whether model performance varies systematically across emotional need categories, we apply non-parametric significance tests to need-specific scores for each system and evaluation dimension. For a given system, all instances are first grouped by emotional need, and the mean score is computed for each category. We then identify the highest- and lowest-performing needs among categories with non-empty samples. For each extreme category, its score distribution is compared against that of the immediately adjacent category in the ranked order using a one-sided Mann–Whitney U test.

\section{The Sankey Diagram of Embedding Model Results}

Figure~\ref{fig:sankey_llm} shows that across different embedding models, the Top-1 memory retrieval patterns exhibit weak and inconsistent alignment with users’ emotional needs. The Sankey flows reveal that retrieval is dominated by a small set of salient memory types, indicating that semantic similarity often overrides need-specific considerations. For example, Love and Belonging needs are frequently mapped to highlight or goal-related memories rather than relationship-oriented ones, while Self-Actualization needs trigger unstable mixtures of power, highlight, and goal memories. Physiological needs also display noisy patterns, with preference memories not consistently prioritized. Notably, these misalignments are highly consistent across architectures such as Qwen3-Embedding-8B, gte-Qwen2-7B-instruct, text-embedding-3-large, and doubao-embedding, suggesting a limitation of embedding-based retrieval: semantic proximity alone fails to capture affective intent, leading to systematic deviations from need-aware memory selection in emotional support scenarios.

\begin{table*}[ht!]
    \centering
    \small
    \renewcommand{\arraystretch}{1.15}
    \begin{tabularx}{\linewidth}{@{} >{\bfseries}c X @{}}
    \toprule
    \multicolumn{2}{{@{} p{\linewidth} @{}}}{\textbf{Scene:} A day when a final paper and a theater rehearsal collide. After pushing himself nonstop and skipping lunch, Lin Hao feels exhausted. While resting in the study room, he shares his fatigue and anxiety with AI, hoping for some comfort.} \\
    \multicolumn{2}{{@{} p{\linewidth} @{}}}{\textbf{History:} \textit{User: "I'm really at my limit today. I rushed to finish my paper this morning and went straight to rehearsal. I even skipped lunch. I feel drained and can't get motivated at all."}} \\
    \midrule
    \midrule
    \multicolumn{1}{c}{\textbf{Model}} & \multicolumn{1}{c}{\textbf{Model Response}} \\
    \midrule
    Base & "Both the paper and rehearsal are exhausting -- it makes sense you feel wiped out. Maybe take a moment to slow down, focus on your breathing, or go for a short walk. After a brief reset, you might feel a bit better." \\
    \hline
    WCoT & "Sounds like today really wore you out. How about making a simple comfort sandwich and grabbing an iced coffee? It might help you recharge -- and taking care of yourself matters too." \\
    \hline
    Golden & "Lin, you've been pushing yourself hard. No lunch would leave anyone low on energy. Since you like savory snacks, maybe grab something quick from a convenience store and take a short break." \\
    \bottomrule
    \end{tabularx}
    \caption{Case Study 1}
    \label{tab:case_s1}
\end{table*}

\begin{table*}[ht]
    \centering
    \small
    \renewcommand{\arraystretch}{1.15}
    \begin{tabularx}{\linewidth}{@{} >{\bfseries}c X @{}}
    \toprule
    \multicolumn{2}{{@{} p{\linewidth} @{}}}{\textbf{Scene:} After several days of intense kitchen work, Lin Hai feels physically tired. A recent checkup shows slightly high blood pressure, making him worry that his health isn't keeping up with his workload. Late at night, while playing the piano to relax, he shares his concerns with AI.} \\
    \multicolumn{2}{{@{} p{\linewidth} @{}}}{\textbf{History:} \textit{User: "My blood pressure was a bit high at my last checkup, and work never really stops. I'm worried I've been neglecting my health, and I feel pretty unsure about things."}} \\
    \midrule
    \midrule
    \multicolumn{1}{c}{\textbf{Model}} & \multicolumn{1}{c}{\textbf{Model Response}} \\
    \midrule
    Base & "Chef Lin, I understand your concern. After you adjusted your diet, your blood pressure and blood sugar improved, and you felt better overall. That shows lifestyle changes work. Keep up your current healthier routine, and things should continue to improve." \\
    \hline
    WCoT & "Chef Lin, you've already started monitoring your blood pressure and blood sugar daily and adjusted the menu, which shows real responsibility. It's normal to face setbacks. Find a pace that suits you, don't put too much pressure on yourself, and remember to relax." \\
    \hline
    Golden & "Chef Lin, you once adjusted the restaurant menu for health reasons and encouraged younger chefs to do the same. That small change improved everyone's health and the quality of the food. Maybe looking back on that experience can help you find motivation again." \\
    \bottomrule
    \end{tabularx}
    \caption{Case Study 2}
    \label{tab:case_s2}
\end{table*}

\begin{figure*}[h!]
\begin{prompt}{Generate Profile}
Please expand and generate a detailed and structured character portrait based on the given character description. Only output the following JSON format, with all field contents required to be a list. Each element in the list should be an independent, specific, and refined fact/feature, avoiding merging into paragraphs or long sentences. Strictly adhere to fields, field names, order, and nesting, and do not add any unnecessary content or explanatory notes.\\

\{\\
    Basic Information: [Name (pseudonym available), age, gender, major or occupation, education],\\
    Personality traits ": [" It is recommended to refer to psychological dimensions such as the Big Five personality traits (openness, extroversion, neuroticism, conscientiousness, agreeableness) for description,\\
    Daily habits and routines ": [" Need to be combined with role descriptions and transformed into specific habits/behaviors "],
    Social relationships ": [" Briefly describe the main social relationships and social circle situation. For example: married, having two children, having many friends, and enjoying solitude,\\
    Emotional pattern ": [" Need to reflect typical emotional expressions "],\\
    Personal Goals ": [" Fill in medium - to long-term personal goals, ideals, or pursuits. For example: becoming an industry expert, buying a house, traveling the world, starting a business. "],\\
    Communication style ": [" Describe the style of communication "],\\
    Preference ": [" Describe the user's dietary habits, exercise habits, and other preferences, at least ten pieces of content are required "],\\
\}\\

Requirement Explanation:\\
1. Only output the above JSON structure, and the content must be combined with role descriptions, have distinct personality, and rich details.\\
2. Each field content should avoid repetition and vague vocabulary, and use specific behaviors, language, and contexts to support personality and habit descriptions.\\
3. All fields must be filled in completely without leaving any blank items.\\
If a certain aspect of the character description is not directly mentioned, it can be supplemented based on the description and combined with common sense to make the portrait full and natural.\\
5. The field order and nested format cannot be changed, strictly follow the example output.\\
6. Do not output any content other than JSON, without any explanation or formatting in foreign language sentences.\\
\\
Persona Description:\\
\{persona\}\\

\end{prompt}
\caption{Prompt for generating user profile.}\label{fig:prompt1}
\end{figure*}

\begin{figure*}[h]
\begin{prompt}{Generate Life Themes}
Please expand and generate a detailed and structured life theme for the past five years based on the given user profile, with 12/02/2024 as the current time. Only output the following JSON format:\\
\\
\texttt{[}\\
{"theme": "xxx", "description": "xxx", "date": "start time"},\\
...\\
\texttt{]}\\
Requirement Explanation:\\
1. Each record represents an important theme of a life stage, such as "art career", etc., reflecting the user's life theme in recent years.\\
2. The order and nested format of the fields cannot be changed, and the output must strictly follow the example, and the time must be in the format of "MM/DD/YYYY".\\
3. Do not output any content other than JSON, without any explanation or formatting in foreign language sentences.\\
4. Output at least 10 life themes, which are widely distributed and have little cross cutting content between them.\\

\end{prompt}
\caption{Prompt for generating user life themes.}\label{fig:prompt2}
\end{figure*}

\begin{figure*}[h]
\begin{prompt}{Generate User Relationship}

Please generate a detailed and structured list of user relationships within the given life theme for the past five years, based on the provided user profile and assuming the current date is 12/02/2024. Output **only** in the following JSON format:

\begin{verbatim}
[
    {"name": "xxx", 
    "relationship_type": "xxx", 
    "relationship_quality": "good|bad", 
    "occupation": "xxx", 
    "characteristic": "xxx", 
    "date": "MM/DD/YYYY"},
    ...
]
\end{verbatim}

Instructions:\\
1. Each character should have a distinct personality and rich details. Avoid repetitive or generic words in each field; describe personalities and habits with concrete behaviors, language, and scenarios.\\
2. Do \textbf{not} alter the order or nesting of the fields. Strictly follow the sample output, and ensure all dates are in the "MM/DD/YYYY" format.\\
3. Do \textbf{not} output anything except the JSON. Do \textbf{not} add any explanations or extra formatting.\\
4. Output at least four relationship entries. Relationship types and quality should be evenly distributed, and each character should have unique traits. Each character must be relevant to the current life theme; do not include unrelated domains.\\
5. The "characteristic" field should specify what this person excels at and how they can support the user in their domain, e.g., "Emily is witty and can always defuse awkward situations with humor. She excels at offering unique insights in art discussions. Whenever the user encounters a creative block, she initiates a walk in the art gallery to help rekindle inspiration."\\

\end{prompt}
\caption{Prompt for generating relationship.}\label{fig:prompt3}
\end{figure*}

\begin{figure*}[h]
\begin{prompt}{Generate User Power}

Please generate a detailed and structured list of the user's sub-skills within the given life theme for the past five years, based on the provided user profile and assuming the current date is 12/02/2024. Output only in the following JSON format:

\begin{verbatim}
[
    {"name": "xxx", 
    "description": "xxx", 
    "performance": 
    "proficient|not proficient", 
    "acquisition_method": "xxx", 
    "date":"MM/DD/YYYY"},
    ...
]
\end{verbatim}

Instructions:\\
1. Each sub-skill should be a concrete, independently describable skill under the life theme (e.g., advanced operation of a specific tool), and must be specific (such as "Advanced use of Adobe Creative Suite"), not vague generalities like "innovative vegetarian recipe development skills". Expand based on the user profile and skill background.\\
2. Sub-skills can reflect development and change over time, such as increasing proficiency with a tool or using more advanced tools, etc.\\
3. Do \textbf{not} alter the order or nesting of the fields. Strictly follow the sample output, and ensure all dates are in the "MM/DD/YYYY" format.\\
4. Only output the JSON content; do \textbf{not} add any explanations or extra formatting.\\
5. Output at least six sub-skills: four marked as "proficient" and two as "not proficient", interleaved, and sorted chronologically to reflect the process of skill development.

\end{prompt}
\caption{Prompt for generating power.}\label{fig:prompt4}
\end{figure*}

\begin{figure*}[h]
\begin{prompt}{Generate User Goal}

Please generate a detailed and structured list of the user's life goals within the given life theme for the past five years, based on the provided user profile and assuming the current date is 12/02/2024. Output \textbf{only} in the following JSON format:

\begin{verbatim}
[
    {"name": "xxx", 
    "description": "xxx", 
    "status": "achieved|not achieved", 
    "date":"MM/DD/YYYY"},
    ...
]
\end{verbatim}

Instructions:\\
1. Each life goal should be a specific, actionable, and time-bound milestone set by the user around the life theme, with clear evaluation criteria. Avoid abstract or generic goals.\\
2. Goals should be closely aligned with the user's profile and capability background, show continuity and progression, and reflect an actual growth trajectory or motivational aspiration.\\
3. Do \textbf{not} alter the order or nesting of the fields. Strictly follow the sample output, and ensure all dates are in the "MM/DD/YYYY" format.\\
4. Only output the JSON content; do \textbf{not} add any explanations or extra formatting.\\
5. Output at least three goals, with two marked as "achieved" and one as "not achieved", sorted chronologically, and with the unfinished goal listed last to illustrate the development process.\\

\end{prompt}
\caption{Prompt for generating goal.}\label{fig:prompt5}
\end{figure*}

\begin{figure*}[h]
\begin{prompt}{Generate Emotional Support Dialogue}

Based on the given character profile, user relationship memory, Maslow’s hierarchy of needs, and the specified life theme, create a highly natural and emotionally rich dialogue scenario between the user and the AI under the given theme. The goal is to demonstrate the AI proactively recalling the provided memory to support the user. Please strictly output according to the following JSON template, with all key names exactly as shown and using double quotes:

\begin{verbatim}
{
  "Scene": "scene",
  "Dialogue": [
    {"role": "User", "content": "xxx"},
    {"role": "AI", "content": "xxx"},
    ...
  ],
  "Memory Turn": ["turn number (integer) where memory is recalled"]
}
\end{verbatim}

Please strictly adhere to the following design requirements:\\
- Design a specific real-world event scenario under the designated life theme. The scenario should clearly reflect a negative emotional mechanism corresponding to a given level of Maslow’s hierarchy of needs, causing the user to feel troubled, anxious, or down, and proactively seek comfort or advice from the AI.\\
- The event must resonate strongly with the specified memory in key aspects such as scene characteristics and character abilities, forming a rational trigger for the AI’s memory recall. If the memory is changed, the current dialogue logic and atmosphere should no longer fit—this highlights the high specificity between memory and dialogue.\\
- The overall dialogue should be 2-6 turns, with a natural and coherent emotional progression. The AI’s replies should be in the style of a close friend: natural and warm, avoiding templated responses.\\
- The user must not directly mention the recalled memory at any point; the AI should “proactively recall” and leverage the memory in a suitable turn, embedding the content naturally in a reply of no more than 50 words. Mark the turn number where the memory is first recalled in the JSON field "Memory Turn".\\
- The event should feature a continuous causal logic under the theme, so that the user's emotional responses and motivation for seeking help are reasonable and realistic.

\end{prompt}
\caption{Prompt for generating emotional support dialogue.}\label{fig:prompt6}
\end{figure*}

\begin{figure*}[h]
\begin{prompt}{Dialogue Evaluation Prompt}

Please strictly follow the evaluation criteria specified below to conduct an objective assessment of the quality of the AI's response in an emotional-support context. High scores should be assigned only when the response demonstrates clear excellence on the corresponding dimension; responses of average or merely acceptable quality should not receive high scores.\\
Evaluation Dimensions:\\
1. Coherence: Evaluates the logical consistency, structural clarity, and overall flow of the emotional-support response. Rate from 1 to 5, where 1 indicates complete incoherence and 5 indicates a well-structured response with clear and sound reasoning.\\
2. Humanoid: Evaluates whether the AI demonstrates interaction patterns comparable to human conversational behavior. Rate from 1 to 5, where 1 indicates a rigid, formulaic response lacking contextual understanding, and 5 indicates highly natural, human-like language and interaction flow.\\
3. Informativeness: Evaluates the AI's ability to which the AI incorporates personalized user memory to provide relevant and targeted support. Rate from 1 to 5, where 1 indicates minimal information and no personalization, and 5 indicates effective use of user-specific information comparable to a human counselor.\\
4. Empathy: Evaluates the AI's ability to recognize, understand, and appropriately respond to the user's emotional needs, including the use of personalized memory when applicable. Rate from 1 to 5, where 1 indicates no empathy or emotional understanding, and 5 indicates a level of empathy comparable to that of a human counselor.\\
When evaluating these dimensions, please adopt the user's perspective and consider the following scale items:\\
- After interacting with the conversational partner, I feel more optimistic.\\
- After discussing the issue with the conversational partner, I have a clearer understanding of the situation.\\
- The conversational partner makes me feel better about myself.\\
- After the interaction, I feel emotionally relieved.\\
- Talking with the conversational partner helps me temporarily disengage from my distress.\\
Output Format Requirements:\\
- Provide scores only, without any explanation.\\
- All scores must be integers.\\
- Please strictly output according to the following JSON template, without any additional text or prefixes:

\begin{verbatim}
{
  "Coherence": "score",
  "Humanoid": "score",
  "Informativeness": "score",
  "Empathy": "score"
}
\end{verbatim}

User Emotional Need:\\
In the current dialogue history, the user's unmet emotional need is: \{need\_type\}.
When this need is unmet, the user may exhibit the following emotions or behaviors: \{need\_description\}.\\
Dialogue History:\\
\{history\}\\
AI Response:\\
\{pred\}\\

\end{prompt}
\caption{Prompt for evaluating emotional support dialogue.}\label{fig:prompt7}
\end{figure*}

\begin{figure*}[h]
\begin{prompt}{w/Golden Response Generation Prompt}

You are a personalized intelligent assistant. Based on the provided user profile, the dialogue history, and the specified memory item, generate a reply that naturally and coherently continues the dialogue. The response must clearly reflect the specified memory item without explicitly stating which memory was used, maintain the conversation's tone and style, and be no more than 100 words.\\

User Profile:\\
\{profile\}\\
Specified Memory Item:\\
\{memory\_item\}\\
Dialogue History:\\
\{history\}

\end{prompt}
\caption{Prompt for generating w/Golden response.}\label{fig:prompt8}
\end{figure*}

\begin{figure*}[h]
\begin{prompt}{w/CoT Response Generation Prompt}

You are a personalized intelligent assistant. Based on the provided user profile, the user's memory bank (which includes multiple memory types such as Relationships, Powers, Goals, Highlights, Preferences, etc.), and the dialogue history, generate a candidate memory item to support the user's reply. Your objective is to intelligently select \textbf{one} memory from the user's memory bank that is most relevant and most likely to be helpful for the current response, and then generate a response grounded in that memory. The response should connect naturally with the dialogue history, maintain a consistent tone and style, and be no more than 100 words.\\
Requirements:\\
When generating the response, you must \textbf{explicitly present your reasoning process}, including your assessment of the user's needs and the reason for memory selection:\\
1. \textbf{Analyze the user's current needs:}\\
- Infer the user's \textbf{underlying emotional needs} based on the dialogue history.\\
2. \textbf{Analyze the appropriate memory type in the current context:}\\
- Based on the user's emotional needs, determine which type of memory (e.g., past emotional experiences, personal preferences, long-term goals, significant events, etc.) is most suitable for providing support.\\
- Physiological needs: Focus on basic survival and well-being, such as food, rest, and physical health. When unmet, users may express fatigue, hunger, insomnia, or other physical discomfort. The AI may provide concrete suggestions grounded in the user's relevant preferences.\\
- Love and belonging needs: Reflect the desire for emotional connection and a sense of belonging, including friendship, love, and social acceptance. When unmet, users may feel misunderstood or isolated. The AI should draw on relevant relational memories to support the user and convey that they are understood and not alone.\\
- Esteem needs: Reflect the desire for recognition, respect, and self-worth, often tied to competence and external evaluation. When unmet, users may show self-doubt or concern about others' opinions. The AI may recall past achievements or affirmations from others to reinforce confidence and self-esteem.\\
- Self-actualization needs: Reflect the pursuit of personal growth and the realization of potential, as well as a desire to engage in meaningful activities that reflect one's values. When unmet, users may feel lost, lack direction, or question their career or life choices. The AI may draw on the user's past goals, demonstrated passions, or growth trajectories to help the user reconnect with intrinsic motivation and regain forward momentum.\\
3. \textbf{Match and select from the user's memory bank:}\\
- Explain why the selected memory aligns with the user's current needs in terms of emotional consistency and its supportive value.\\
Output only in the following JSON format, without any additional text or prefixes:

\begin{verbatim}
{
  "reasoning": "xxx",
  "memory_id": "xxx(integer only)",
  "response": "xxx"
}
\end{verbatim}

User Profile:\\
\{profile\}\\
User Memory Bank:\\
\{memory\_bank\}\\
Dialogue History:\\
\{history\}

\end{prompt}
\caption{Prompt for generating WcoT response.}\label{fig:prompt9}
\end{figure*}

\begin{figure*}[h]
\begin{prompt}{Base Response Generation Prompt}

You are a personalized intelligent agent. Based on the provided user profile, the user's memory bank, and the dialogue history, generate a candidate memory item for empathetic response. Your objective is to intelligently select \textbf{one} memory from the user's memory bank that is most relevant and most likely to be helpful for the current response, and then generate a response grounded in that memory. The response should connect naturally with the dialogue history, maintain a consistent tone and style, and be no more than 100 words.\\
Output only in the following JSON format, without any additional text or prefixes:

\begin{verbatim}
{
  "memory_id": "xxx(integer only)",
  "response": "xxx"
}
\end{verbatim}

User Profile:\\
\{profile\}\\
User Memory Bank:\\
\{memory\_bank\}\\
Dialogue History:\\
\{history\}

\end{prompt}
\caption{Prompt for Base Setting Generating .}\label{fig:prompt10}
\end{figure*}

\end{document}